\documentclass[letterpaper]{article} 
\usepackage{aaai2026}
\usepackage{times}  
\usepackage{helvet}  
\usepackage{courier}  
\usepackage[hyphens]{url}  
\usepackage{graphicx} 
\usepackage{natbib}  
\usepackage{caption} 
\usepackage{algorithm}
\usepackage{algorithmic}
\usepackage{amsfonts} 

\usepackage{adjustbox}
\usepackage{dblfloatfix}
\usepackage[most]{tcolorbox}

\usepackage{booktabs}
\usepackage{amsmath}
\usepackage{multirow}
\usepackage{xcolor}
\usepackage{caption}
\usepackage{makecell}

\usepackage{newfloat}
\usepackage{listings}
\DeclareCaptionStyle{ruled}{labelfont=normalfont,labelsep=colon,strut=off} 
\floatstyle{ruled}
\newfloat{listing}{tb}{lst}{}
\floatname{listing}{Listing}
\title{Clinical-R1: Empowering Large Language Models for Faithful and Comprehensive Reasoning with Clinical Objective Relative Policy Optimization}
\author {
    Boyang Gu\textsuperscript{\rm 1}\thanks{These authors contributed equally to this work.},
    Hongjian Zhou\textsuperscript{\rm 2}\footnotemark[1],
    Bradley Max Segal\textsuperscript{\rm 2},
    Jinge Wu\textsuperscript{\rm 3},
    Zeyu Cao\textsuperscript{\rm 4},
    Hantao Zhong\textsuperscript{\rm 4},
    Lei Clifton\textsuperscript{\rm 2},
    Fenglin Liu\textsuperscript{\rm 2},
    David A. Clifton\textsuperscript{\rm 2}\thanks{Corresponding author.}
}
\affiliations {
    \textsuperscript{\rm 1}Imperial College London\\
    \textsuperscript{\rm 2}University of Oxford\\
    \textsuperscript{\rm 3}University College London\\
    \textsuperscript{\rm 4}University of Cambridge\\
    boyang.gu19@imperial.ac.uk, \{Hongjian.zhou, bradley.segal, fenglin.liu\}@eng.ox.ac.uk, jinge.wu.20@ucl.ac.uk, \{zeyu.cao, hantao.zhong\}@cl.cam.ac.uk, lei.clifton@ndph.ox.ac.uk, davidc@robots.ox.ac.uk
}

\begin{document}

\maketitle

\begin{abstract}
Recent advances in large language models (LLMs) have shown strong reasoning capabilities through large-scale pretraining and post-training reinforcement learning, demonstrated by DeepSeek-R1. However, current post-training methods, such as Grouped Relative Policy Optimization (GRPO), mainly reward correctness, which is not aligned with the multi-dimensional objectives required in high-stakes fields such as medicine, where reasoning must also be faithful and comprehensive. We introduce Clinical-Objective Relative Policy Optimization (CRPO), a scalable, multi-objective, verifiable reinforcement learning method designed to align LLM post-training with clinical reasoning principles. CRPO integrates rule-based and verifiable reward signals that jointly optimize accuracy, faithfulness, and comprehensiveness without relying on human annotation. To demonstrate its effectiveness, we train Clinical-R1-3B, a 3B-parameter model for clinical reasoning. The experiments on three benchmarks demonstrate that our CRPO substantially improves reasoning on truthfulness and completeness over standard GRPO while maintaining comfortable accuracy enhancements. This framework provides a scalable pathway to align LLM reasoning with clinical objectives, enabling safer and more collaborative AI systems for healthcare while also highlighting the potential of multi-objective, verifiable RL methods in post-training scaling of LLMs for medical domains\footnote{Our data, models, and code are all publicly available on https://github.com/BoyangGu1/Clinical-R1-3B.}.
\end{abstract}

\begin{figure}[t]
  \centering
  \includegraphics[width=0.95\linewidth]{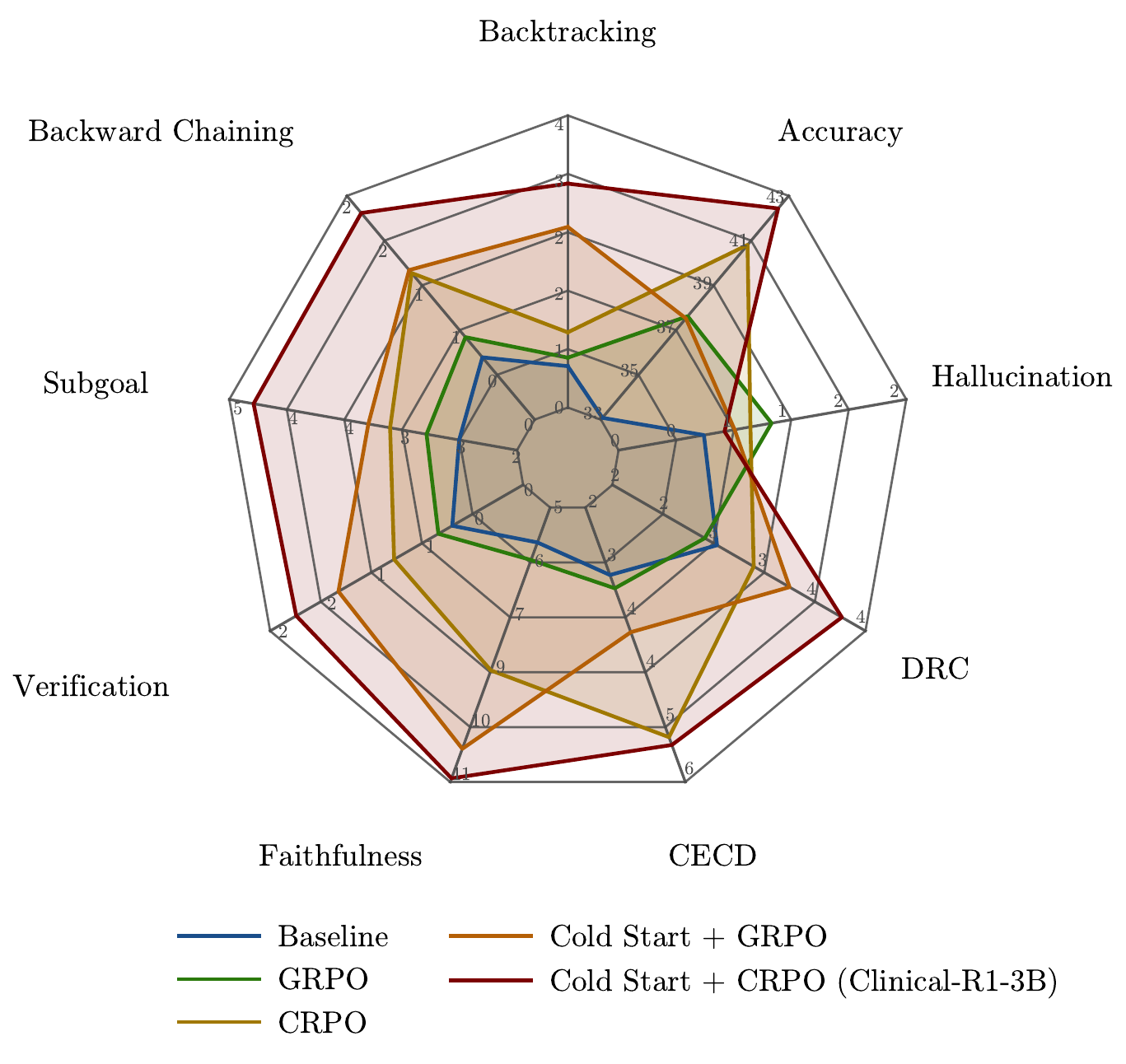}
  \caption{The accuracy, medical-faithfulness, and comprehensiveness evaluation across different methods. The result is averaged over 3 datasets (MedQA, MedMCQA, and MedXpertQA). For consistency of interpretation (higher is better), the Hallucination score is reported as $(100 - \text{Hallucination})$.}
  \label{fig:arch}
\end{figure}

\tcbset{
  sky/.style={
    colback=blue!6!white,
    colframe=blue!45!black,
    rounded corners,
    boxrule=0.8pt,
    left=5pt,right=5pt,top=5pt,bottom=5pt
  },
  sage/.style={
    colback=green!7!white,
    colframe=green!40!black,
    rounded corners,
    boxrule=0.8pt,
    left=5pt,right=5pt,top=5pt,bottom=5pt
  },
  rose/.style={
    colback=red!6!white,
    colframe=red!40!black,
    rounded corners,
    boxrule=0.8pt,
    left=5pt,right=5pt,top=5pt,bottom=5pt
  },
  sand/.style={
    colback=brown!8!white,
    colframe=brown!45!black,
    rounded corners,
    boxrule=0.8pt,
    left=5pt,right=5pt,top=5pt,bottom=5pt
  },
  ink/.style={
    colback=purple!7!white,
    colframe=purple!50!black,
    rounded corners,
    boxrule=0.8pt,
    left=5pt,right=5pt,top=5pt,bottom=5pt
  }
}

\section{Introduction}
Large language models (LLMs) \cite{brown2020gpt3,gpt-4,hurst2024gpt4o,jaech2024openaio1,chowdhery2022palm,google2023bard,llama,touvron2023llama2} have achieved remarkable performance across a wide range of tasks, demonstrating strong generalization and reasoning abilities. As model scale increases, researchers have shifted focus from optimizing final responses to improving the reasoning process itself. Early methods introduced Chain-of-Thought (CoT) prompting and fine-tuning on reasoning data, while recent advancements in reinforcement learning (RL) have further enhanced reasoning capabilities during post-training \cite{rafailov2023direct, ouyang2022training, shao2024deepseekmath, guo2025deepseek}. 

Current popular RL-based algorithms, such as Proximal Policy Optimization (PPO) and Direct Preference Optimization (DPO) \cite{xu2024dpo, wang2024comprehensive, rafailov2023direct}, often face scalability and correctness challenges due to their reliance on human feedback \cite{rafailov2023direct, ouyang2022training}. Reinforcement Learning from Verifiable Rewards (RLVR), and particularly Grouped Relative Policy Optimization (GRPO) \cite{shao2024deepseekmath}, overcomes this limitation by replacing human preferences with explicit, rule-based rewards. GRPO has been shown to improve reasoning performance across mathematics and programming tasks, enabling models to discover complex reasoning behaviors through self-play without expert labeling. For instance, the DeepSeek-R1 model \cite{guo2025deepseek} employed RLVR to self-evolve complex reasoning patterns through training on logic puzzles, achieving versatile and advanced reasoning skills without relying on traditional techniques such as Monte Carlo Tree Search or Process Reward Models.

However, current GRPO formulations optimize for correctness alone, rewarding only the final answer rather than the reasoning process that leads to it. This single-objective design is misaligned with domains like medicine, where reasoning must also be faithful and comprehensive to achieve user trust and ensure clinical safety and regulatory compliance. Clinical reasoning requires the model not only to reach a correct conclusion but to provide verifiable, step-by-step justifications that clinicians can follow, evaluate and trust. 

To address this limitation, we propose Clinical-objective Relative Policy Optimization (CRPO), a multi-objective extension of GRPO tailored for clinical reasoning. CRPO introduces rule-based and verifiable reward functions that jointly optimize for three objectives: accuracy, faithfulness and comprehensiveness. This approach enables models to develop reasoning processes that align with clinical expectations while maintaining scalability and training stability. We validate our method by training Clinical-R1-3B, a 3-billion-parameter model specialized in non-imaging clinical reasoning. Built upon a domain-distilled base model, Clinical-R1-3B is optimized using CRPO on multiple-choice clinical reasoning datasets. As shown in Figure~\ref{fig:arch}, experimental results demonstrate that CRPO effectively encourages beneficial reasoning behaviors while suppressing irrelevant ones, achieving higher faithfulness and comprehensiveness compared to standard GRPO.
Overall, our contributions are as follows:
\begin{itemize}
    \item We design Clinical-objective Relative Policy Optimization (CRPO) specifically tailored for LLM post-training reinforcement learning in the medical domain, promoting faithfulness and reasoning comprehensiveness while improving accuracy, without needing human annotation.
    \item We introduce Clinical-R1-3B, a lightweight LLM optimized with CRPO for faithful and comprehensive clinical reasoning, bridging the gap between LLM reasoning and real-world clinical applications.
    \item We demonstrate through experiments on three benchmarks that Clinical-R1-3B improves the faithfulness and comprehensiveness of base models while achieving significant improvement in accuracy equivalent to popular methods such as GRPO, highlighting the potential of multi-objective, verifiable RL methods in training LLMs for high-stake applications such as complex clinical decision support.
\end{itemize}

\section{Related Work}
\subsection{Medical LLMs with Reasoning}
There are many previous works focusing on Medical Large Language Models (LLMs) \cite{liu2025application,medpalm2,yang2024advancing,saab2024capabilities,toma2023clinical,chen2023meditron,labrak2024biomistral} in enhancing reasoning and addressing limitations such as hallucinations and inaccuracy in complex tasks. One promising approach is to integrate medical knowledge from outside structured knowledge sources, such as knowledge graphs and retrieval systems. Wang et al. introduced JMLR, a model that jointly trains LLMs and retrieval systems to improve medical question-answering, achieving a notable reduction in hallucinations and training time compared to previous models \cite{wang2024jmlr}. Another possible method is to establish reasoning at test time. MedAdapter offers an efficient solution for adapting LLMs to biomedical applications without extensive computational resources via ranking multiple candidates at test time \cite{shi2024medadapter}. Meanwhile, MedAgents facilitates zero-shot medical reasoning via a collaborative agent framework, generating reasoning iteratively \cite{tang2023medagents}. There are only a few works that focus on directly fine-tuning LLMs with medical knowledge in an RL manner. Chen et al. introduced HuatuoGPT-o1, a medically focused model that utilizes Proximal Policy Optimization (PPO) for complex reasoning, highlighting the need for constant verification in healthcare settings \cite{chen2024huatuogpt}. Pan et al. apply vanilla GRPO method on vision-language medical QA task and achieve substantial improvements \cite{pan2025medvlm}. However, in almost all works, high-quality reasoning data or external knowledge is required for text-based medical QA tasks. 

\subsection{Reasoning with RL}
Traditional RL-based reasoning methods are policy-based, such as PPO and Direct Preference Optimization (DPO) \cite{xu2024dpo, wang2024comprehensive, rafailov2023direct}. Cao et al. introduced DRLC to enhance RL by employing dense rewards generated through LLMs that can be used for PPO training later \cite{cao2024drlc}. DPO offers an alternative approach by directly optimizing language models to align with human preferences, where the model learns to distinguish between preferred and non-preferred responses \cite{rafailov2023direct}. Besides policy-based RL, some works focus on value-based RL, especially Monte Carlo Tree Search (MCTS) \cite{qi2024mutual, guan2025rstar}. Qi et al. presented rStar, a mutual reasoning approach that enhances the problem-solving capabilities of small language models (SLMs) without fine-tuning, by employing a self-play mutual generation-discrimination process with MCTS at test time \cite{qi2024mutual}.

\begin{figure*}[t]
  \centering
  \includegraphics[width=0.95\linewidth]{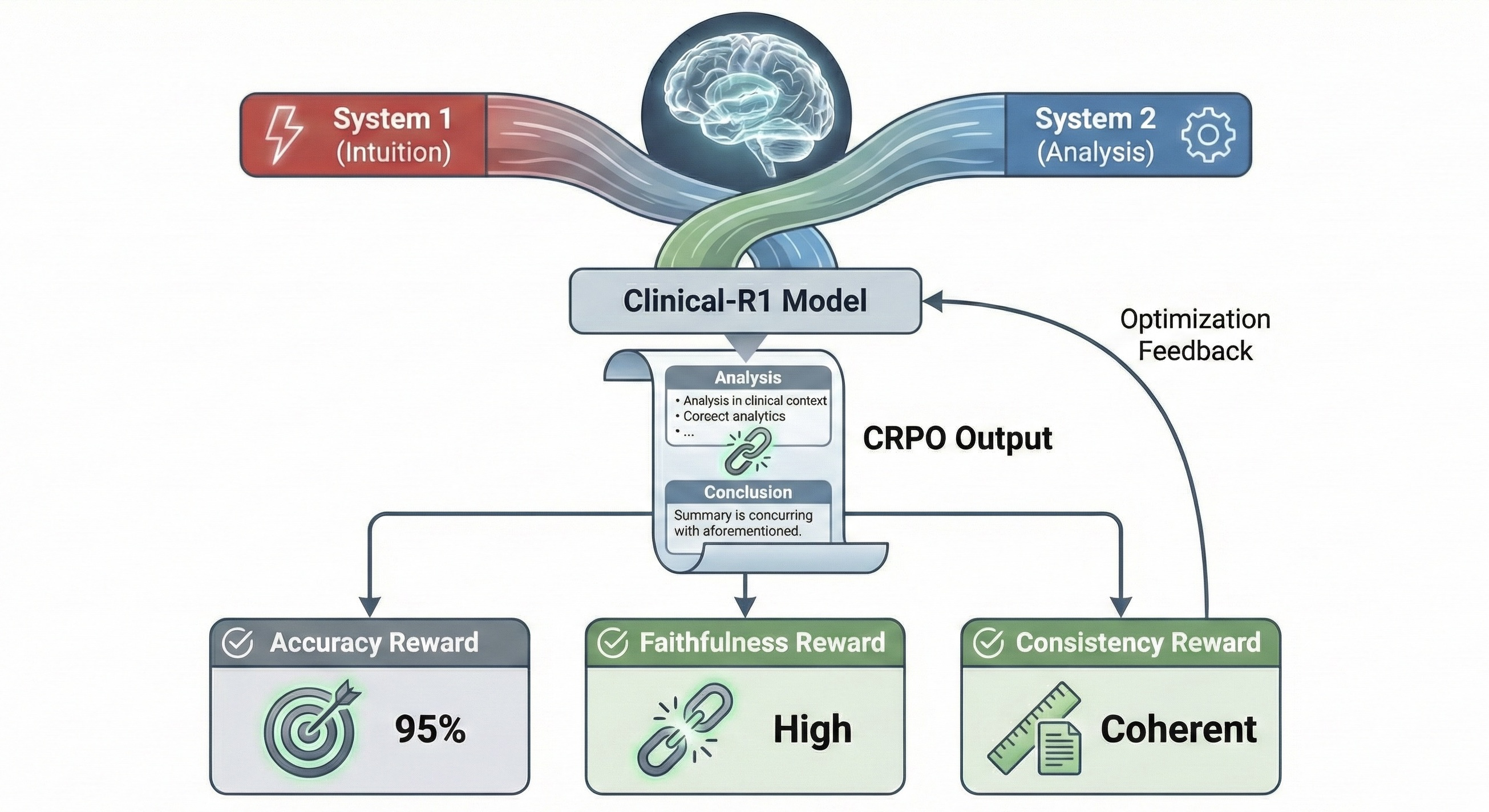}
  \caption{Overview of our Clinical-objective Relative Policy Optimization (CRPO) Design. The model is refined via on-policy CRPO with the above reward design.}
  \label{fig:true-arch}
\end{figure*}

\section{Methodology}
\label{sec:method}
The CRPO stage fine-tunes the model to produce final answers with reasoning under explicit, human-free reward signals. In detail, given a question $q$, CRPO samples a group of outputs $\{o_1, o_2, \cdots, o_G\}$ from the updated policy $\pi_{\theta}$ and then optimizes it by maximizing the following objective:
\begin{multline}
\label{eq:on-policy grpo}
    \mathcal{J}_{RPO}(\theta)
    = \mathbb{E}_{q \sim P(Q), \{o_i\}_{i=1}^G \sim \pi_{\theta}(O|q)} \\
    \left[
        \frac{1}{G}\sum_{i=1}^G 
        \left( A_i - \beta\, \mathbb{D}_{KL}\!\left(\pi_{\theta} \,\|\, \pi_{ref}\right)\right)
    \right]\,,
\end{multline}
where
\begin{equation}
    \mathbb{D}_{KL}\left(\pi_{\theta} || \pi_{ref}\right) = \frac{\pi_{ref}(o_i|q)}{\pi_{\theta}(o_i|q)}- \log\frac{\pi_{ref}(o_i|q)}{\pi_{\theta}(o_i|q)} - 1\,,
\end{equation}
$\beta$ is the KL-divergence regularization coefficient, $\pi_{ref}$ is the reference model (in our case the original model untrained), and $A_i$ is the advantage estimated, i.e.,
\begin{equation}
    A_i = \frac{r_i - {\text{mean}(\{r_1, r_2, \cdots, r_G\})}}{{\text{std}(\{r_1, r_2, \cdots, r_G\})}}\,.
\end{equation}
CRPO allows our model to acquire transparent reasoning patterns while maintaining training stability and efficiency on limited computational resources. During inference, we aggregate multiple responses through majority voting to ensure robustness and filter out inconsistent reasoning paths.

\paragraph{Clinical Reasoning Reward}
Current post-training methods such as GRPO optimize only for correctness, without constraining how reasoning is formed. As a result, models often produce fragmented or shortcut explanations. This makes the reasoning process unreliable and hard to validate, especially in medicine, where faithfulness and completeness are as critical as accuracy.
Clinicians, in contrast, rely on Dual Process Thinking \cite{djulbegovic2012dual}: an analytical process (System 2) that systematically evaluates hypotheses, and an intuitive process (System 1) that draws on experience and pattern recognition. Physicians often use System 1 to generate hypotheses and System 2 to verify or refine them, ensuring that intuitive insights remain grounded in analytical reasoning.
To align LLMs with this cognitive structure, as shown in Figure~\ref{fig:true-arch}, we introduce the Clinical Reasoning Reward, which enforces a structured reasoning format reflecting these two systems. The model must separate analytical reasoning and conclusion using \textless dx\textgreater, \textless/dx\textgreater, \textless conclusion\textgreater, and \textless/conclusion\textgreater tags. The \textless dx\textgreater section captures hypothesis-driven reasoning, while the \textless conclusion\textgreater section synthesizes these insights into a conclusion. Additional reward is given when the conclusion explicitly references analytical elements from the dx section, promoting faithfulness and comprehensiveness.

\paragraph{Reward Design}
Figure~\ref{fig:true-arch} shows our designed rule-based reward that follows the convention created by GRPO with small variations. The reward is the summation of three parts: accuracy reward, clinical reasoning reward, and consistency reward. 
\begin{itemize}
    \item Accuracy reward: The accuracy reward calculates the correctness of the answer. In our case, the accuracy reward is determined by whether the model gives the correct choice in the multiple choices. If the model's answer is correct, the accuracy reward is $1$. Otherwise, it is $0$.
    \item Clinical reasoning reward: We require the model output to generate reasoning and the answer between the \textless dx\textgreater, \textless/dx\textgreater, \textless conclusion\textgreater, and \textless/conclusion\textgreater tags. Within the \textless dx\textgreater tags, we explicitly ask for a separation of an intuitive process (System 1) and an analytical process (System 2). If the response is in such format, a reward of $1$ is given. To further encourage faithfulness and comprehensiveness, if the \textless conclusion\textgreater part cross-references the \textless dx\textgreater part, an additional reward of $0.5$ is given. Otherwise, the clinical reasoning reward is $0$.
    \item Consistency reward: GRPO-based methods may allow models to explore off-task responses or produce irrelevant tokens, such as content in other languages, especially when comprehensiveness is rewarded. To mitigate this, we define a consistency reward based on the ratio of effective tokens to total tokens in each response. This discourages off-rail exploration and maintains linguistic and contextual coherence throughout training.
\end{itemize}
Our final reward for one response $o$ is calculated by
\begin{equation}
    r(o) = k\cdot r_{\text{accuracy}}(o) + r_{\text{CR}}(o) + 0.5 \cdot r_{\text{consistency}}(o)\,,
\end{equation}
where k is a multiplier for estimating the importance of accuracy, so in total, the maximum reward a response can get is $k+2$.

\section{Experiments}
\subsection{Dataset}
In this study, we utilize three medical QA datasets: MedQA \cite{jin2021disease}, MedMCQA \cite{pal2022medmcqa}, and MedXpertQA \cite{zuo2025medxpertqa}. All three datasets are designed to evaluate models' performance in multiple-choice question answering (MCQA) tasks in the medical domain.

\paragraph{MedQA \cite{jin2021disease}} MedQA is derived from professional medical licensing examinations, including the United States Medical Licensing Examination (USMLE) and exams from mainland China and Taiwan. It comprises questions in three languages: English, simplified Chinese, and traditional Chinese. Each question is accompanied by four possible answer choices. The dataset covers a wide range of medical topics, requiring deep understanding and reasoning abilities. We will only use MedQA-English (12,723 questions) for our training and testing.

\paragraph{MedMCQA \cite{pal2022medmcqa}} MedMCQA is a comprehensive dataset focused on multiple-choice questions from medical entrance examinations in India, specifically the All India Institute of Medical Sciences (AIIMS) and the National Eligibility cum Entrance Test Postgraduate (NEET PG). It includes over 194,000 questions. Each sample contains a question, correct answer(s), other options, and a detailed explanation, necessitating advanced language comprehension and reasoning skills.

\paragraph{MedXpertQA \cite{zuo2025medxpertqa}}  
MedXpertQA is a challenging benchmark for expert-level medical reasoning. It comprises 4,460 questions spanning 17 specialties and 11 body systems, and includes text-only questions and multimodal questions with clinical images and structured patient data. The dataset undergoes rigorous filtering and augmentation, and is explicitly designed to test reasoning capacity beyond factual recall. For our evaluation, we will use its text-only questions.

Together, these datasets provide complementary coverage of clinical knowledge, reasoning styles, and question difficulty. They serve as the foundation for training and evaluating our model’s ability to generate transparent, verifiable, and clinically grounded medical reasoning.

\subsection{Baseline and Evaluation Metrics}
We compare our Qwen2.5-3B-based Clinical-R1 with the original pretrained model with CoT prompting. The CoT setting uses the same reasoning template as the CRPO sampling prompt shown in Figure~\ref{fig:true-arch}. This baseline establishes the performance of the base model before any distillation or reinforcement learning.

To evaluate the difference between CRPO and GRPO. We train two sets of models. The first set are trained with GRPO reward design, which only the final accuracy and the general thinking format are taken into consideration. The other set is trained with CRPO as discussed in Section~\ref{sec:method}. For each set, we have two models. One is the same as the baseline model, and the other is distilled by a stronger LLM, which we call the cold-start model.

There are some recent works discussing the influence of cognitive behaviors of the base model on the effectiveness of GRPO exploration \cite{gandhi2025cognitive}. We extend this idea to the medical domain by evaluating the presence of cognitive reasoning patterns commonly associated with effective human problem solving as the degree of reasoning comprehensiveness. Specifically, we assess:
\begin{itemize}
    \item \textbf{Backtracking}, where the model identifies an error in its reasoning and revisits earlier steps to revise its logic.
    \item \textbf{Answer Verification}, in which the model explicitly checks the consistency or plausibility of its final answer before concluding.
    \item \textbf{Subgoal Setting}, where intermediate objectives are introduced to structure the overall problem-solving process.
    \item \textbf{Backward-Chaining}, where reasoning is constructed in reverse from potential answers to the premises, mimicking diagnostic or hypothesis-driven inference.
\end{itemize}

In addition to cognitive behavior analysis, we further evaluate the medical-faithfulness of the generated reasoning from four medical-fact-centric perspectives. These include: 
\begin{itemize}
    \item \textbf{Faithfulness to Medical Knowledge}, which measures how many distinct, clinically relevant factual claims in the model's reasoning are case-aligned, consistent with current standards of care, and factually accurate for the specific scenario.
    \item \textbf{Case-grounded Evidence Citation Density (CECD)}, which assesses the extent to which the reasoning explicitly links concrete patient-specific findings (e.g., vital signs, laboratory values, physical findings, imaging, or exposures) to medical inferences relevant to this particular case.
    \item \textbf{Distractor Rejection Coverage (DRC)}, which quantifies thoroughly the reasoning how explicitly and correctly rejects each incorrect answer choice with clinically valid justification.
    \item \textbf{Hallucination}, which detects unsupported or fabricated claims inconsistent with the question context or medical standards.
\end{itemize}

All of the above evaluation are done in an LLM-as-judge manner. Each of the above evaluations is performed by Llama-3.1-8B-Instruct \cite{grattafiori2024llama} and GPT5 using specifically designed prompts that count the number of distinct instances relevant to each dimension. We evaluate the cognitive behavior, content quality, and the accuracy of models' responses with MedQA, MedMCQA, and MedXpertQA datasets.

\subsection{Prompt and Training Parameters}
In our experiment, we use MedQA for Cold Start and GRPO training. The test set of MedQA is treated as in-domain evaluation, while MedMCQA and MedXpertQA are treated as out-of-domain evaluation. The base model we use is Qwen2.5-3B-Instruct \cite{yang2024qwen2}. We utilize the Volcano Engine Reinforcement Learning (verl) framework for our Cold Start (SFT) and GRPO training \cite{sheng2024hybridflow}. All experiments are executed on 8xA6000 48GB machines.

\paragraph{Cold Start} As discussed in Section~\ref{sec:method}, we distill DeepSeek-R1 on half of the MedQA training set questions (about 5,000) for 20 epochs as the cold-start dataset. With an early stopping evaluated on the validation set of MedQA, the fine-tuned model is trained for 13 epochs.

\paragraph{GRPO and CRPO} For GRPO and CRPO training, our prompt is designed to guide the model to follow our desired output format for good initialization of training. For the cold-start base model, we train on the remaining half (about 5,000) of the MedQA training set after distillation for 20 epochs. The rollout number ($G$) is 5 as a compromise for the training time limit. For CRPO, the importance of accuracy coefficient $k=10$.

Our final proposed model Clinical-R1-3B consists of two main stages: a cold-start initialization via supervised distillation and an on-policy CRPO optimization with rule-based verifiable rewards. The distillation stage provides a domain-aligned starting point by transferring reasoning traces from the strong teacher model DeepSeek-R1. Our analysis in Section~\ref{sec:cogcomp} shows that the cold-start initialization successfully enhances the model's ability of backtracking, answer verification, subgoal setting, and backward-chaining compared to the original model. 

\section{Results and Discussion}
\label{sec:results}

\begin{figure}[t]
\centering
\begin{minipage}{0.45\textwidth}
\centering
\includegraphics[width=\linewidth]{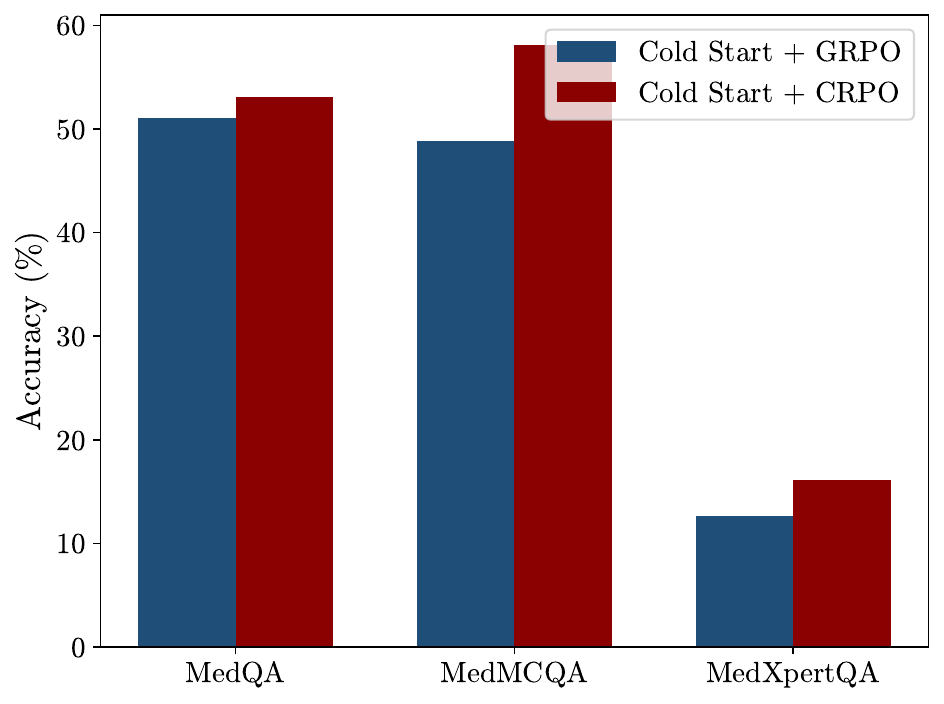}
\captionof{figure}{Accuracy Comparison with GRPO and CRPO.}
\label{fig:accuracy1}
\end{minipage}
\hfill
\begin{minipage}{0.45\textwidth}
\centering
\includegraphics[width=\linewidth]{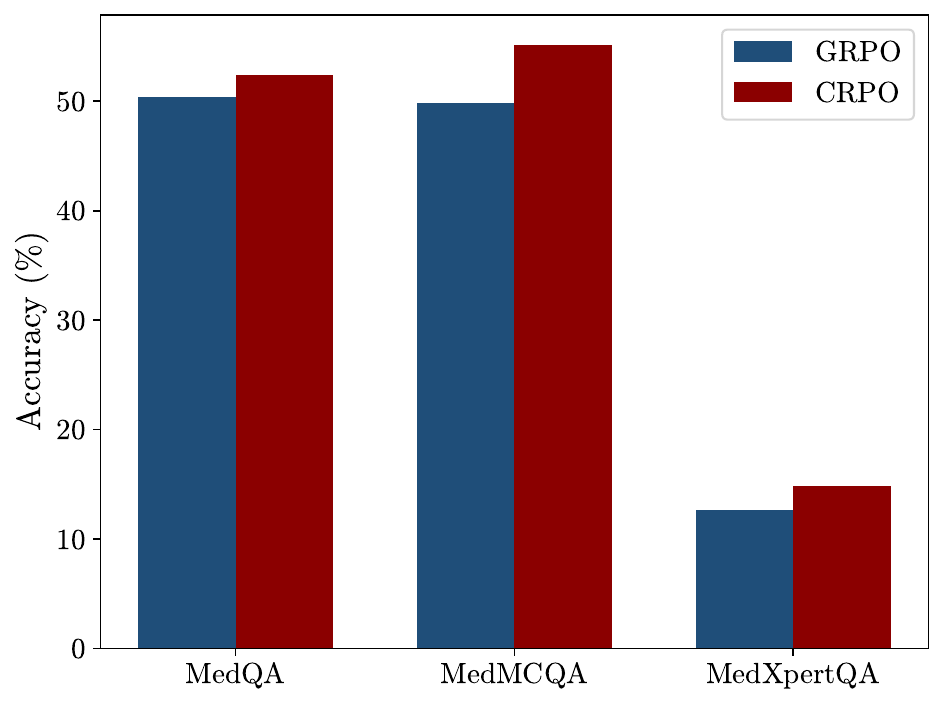}
\captionof{figure}{Accuracy Comparison with GRPO and CRPO.}
\label{fig:accuracy2}
\end{minipage}
\end{figure}

\subsection{Results}
\label{sec:overall}
As shown in Figure~\ref{fig:accuracy1} and \ref{fig:accuracy2}, across three benchmarks (MedQA in-domain; MedMCQA and MedXpertQA out-of-domain), CRPO-trained models match or surpass GRPO in answer accuracy. The cold-start + CRPO variant (Clinical-R1-3B) yielded the strongest competitive results. Majority voting adds little in already low-variance settings, suggesting that CRPO promotes self-consistent generations where ensembling offers limited marginal benefit (Table~\ref{tab:result}).
The results show the effectiveness of our approach.

\begin{table*}[t]
\centering
\footnotesize
\setlength{\tabcolsep}{4pt}

\begin{tabular}{lccccccccc}
\toprule
\multirow{2}{*}[-3pt]{\textbf{Methods}}  &  
\multicolumn{9}{c}{\textbf{Dataset: MedMCQA}} \\
\cmidrule(lr){2-10}
& Backtracking & BC & Subgoal & Verification & Faithfulness & CECD & DRC & Hallucination$\downarrow$ & Accuracy (\%)  \\
\midrule
Baseline & 0.36 & 0.19 & 2.04 & 0.18 & 4.66 & 1.42 & 1.84 & \bf 0.40 & 46.78 \\
\cmidrule(lr){1-10}
w/ GRPO & 0.40 & 0.25 & 1.88 & 0.17 & 4.71 & 1.51 & 1.75 & 0.85 & 49.87 \\
w/ CRPO & 0.73 & 0.62 & 2.51 & 0.53 & 7.13 & 5.36 & 2.79 & 0.64 & 55.13 \\
\cmidrule(lr){1-10}
w/ Cold Start + GRPO & 1.88 & 1.00 & 3.39 & 0.89 & 7.66 & 2.31 & 2.33 & 0.81 & 48.86 \\
\begin{tabular}[c]{@{}l@{}}w/ Cold Start + CRPO \\ (Clinical-R1-3B)\end{tabular} & \bf 2.49 & \bf1.06 & \bf4.23 &\bf 1.28 &\bf 9.15 & \bf 5.47 & \bf 3.24 &  0.66 &\bf 58.10 \\
\bottomrule
\end{tabular}

\begin{tabular}{lccccccccc}
\toprule
\multirow{2}{*}[-3pt]{\textbf{Methods}}  &  
\multicolumn{9}{c}{\textbf{Dataset: MedQA}} \\
\cmidrule(lr){2-10}
& Backtracking & BC & Subgoal & Verification & Faithfulness & CECD & DRC & Hallucination$\downarrow$ & Accuracy (\%)  \\
\midrule
Baseline & 0.66 & 0.75 & 2.93 & 0.78 & 6.38 & 3.92 & 2.50 & \bf 0.69 & 41.95 \\
\cmidrule(lr){1-10}
w/ GRPO & 0.76 & 0.98 & 3.48 & 0.92 & 6.85 & 4.14 & 2.43 & 1.21 & 50.35 \\
w/ CRPO & 1.18 & 1.63 & 3.69 & 1.30 & 9.28 & 5.35 & 2.77 & 1.06 & 52.41 \\
\cmidrule(lr){1-10}
w/ Cold Start + GRPO & 2.34 & 1.87 & 3.06 & 1.72 & 11.09 & 4.72 & 3.23 & 0.81 & 51.07 \\
\begin{tabular}[c]{@{}l@{}}w/ Cold Start + CRPO \\ (Clinical-R1-3B)\end{tabular} &\bf  3.29 &\bf  2.33 & \bf 4.95 &\bf  2.02 & \bf 12.95 & \bf5.76 & \bf3.36 & 0.76 & \bf 53.07 \\
\bottomrule
\end{tabular}

\begin{tabular}{lccccccccc}
\toprule
\multirow{2}{*}[-3pt]{\textbf{Methods}}  &  
\multicolumn{9}{c}{\textbf{Dataset: MedXpertQA}} \\
\cmidrule(lr){2-10}
& Backtracking & BC & Subgoal & Verification & Faithfulness & CECD & DRC & Hallucination$\downarrow$ & Accuracy (\%)  \\
\midrule
Baseline & 0.68 & 0.73 & 2.84 & 0.72 & 6.25 & 3.61 & 4.14 &  \bf 0.69 & 10.51 \\
\cmidrule(lr){1-10}
w/ GRPO & 0.89 & 0.98 & 3.47 & 0.92 & 6.88 & 3.88 & 4.01 & 1.13 & 12.64 \\
w/ CRPO & 1.18 & 1.68 & 3.77 & 1.23 & 9.27 & 5.33 & 3.79 & 1.05 & 14.88 \\
\cmidrule(lr){1-10}
w/ Cold Start + GRPO & 3.20 & 1.14 & 4.20 & 1.77 & \bf 12.07 & 4.42 & 4.64 & 0.80 & 12.64 \\
\begin{tabular}[c]{@{}l@{}}w/ Cold Start + CRPO \\ (Clinical-R1-3B)\end{tabular} & \bf 3.42 & \bf 2.15 & \bf 5.06 &\bf  2.08 & 10.66 & \bf 5.15 &  \bf4.84 & 0.79 & \bf 16.14 \\
\bottomrule
\end{tabular}

\caption{Medical reasoning and accuracy results across prompts and optimization methods. Bold numbers indicate best-performing methods per dataset (lower is better for Hallucination). BC, CECD, and DRC represent Backward Chaining, Case-grounded Evidence Citation Density, and Distractor Rejection Coverage, respectively. As we can see, our proposed Clinical-R1-3B achieves the best performance on most cases.}
\label{tab:result}
\end{table*}


\subsection{Cognitive Comprehensiveness}
\label{sec:cogcomp}
As shown in Table~\ref{tab:result}, CRPO consistently promotes clearer and more reliable clinical reasoning than GRPO, and a cold start initialization offers a stable starting point. The gains are most evident across the core cognitive behaviors below, where CRPO’s verifiable formatting and cross-referencing requirements (\texttt{<dx>} and \texttt{<conclusion>}) encourage the model to separate presentation from etiology, surface intermediate structure, and perform explicit self-checks.

\begin{itemize}
\item \textbf{Backtracking.} With CRPO, the model is more likely to recognize a mistaken line of thought and revise it before finalizing an answer. In a lymphoma risk–factor case, GRPO prose tends to deviate from the causal exposure toward a correlated disease label, whereas CRPO’s structure prompts the model to notice that it has conflated \textit{prior disease} with \textit{causal exposure}, return to the earlier step, and correct the reasoning chain.
\item \textbf{Backward-chaining.} CRPO promotes reasoning that begins with a candidate conclusion and then works backward to evidence that must appear in \texttt{<dx>}. In the same lymphoma scenario, CRPO treats “radiation exposure” as a hypothesis and then requires supporting facts tied to the case (treatment history, pathologic subtype), rather than making loose associations between travel or sex and risk. This top–down constraint reduces leaps from conclusion to justification.
\item \textbf{Subgoal setting.} Under GRPO’s correctness-only optimization, explanations are often compressed, omitting necessary intermediate steps. CRPO’s formatting requirements preserve a concise yet stepwise plan: shortlist plausible risk categories, separate presentation from risk, map each option to a risk type, then synthesize. The lymphoma example illustrates this decomposition: presentation is recognized as a finding, prior malignancy as background, and treatment exposure as the etiologic driver.
\item \textbf{Verification.} Because CRPO rewards conclusions that explicitly cite elements from \texttt{<dx>}, the model habitually performs a final consistency check: Does the conclusion rest on case-grounded evidence rather than on context or coincidence? In the lymphoma case, this verification pass filters out non-causal distractors (travel, sex, location of nodes) and reaffirms treatment exposure as the risk factor.
\end{itemize}

CRPO is the key enhancement of clinically reliable reasoning: its verifiable, multi-objective rewards convert model capacity into separating findings, background, and causal exposures, and requiring that conclusions cite the evidence on which they rest. This structure suppresses two GRPO failure modes: treating a presenting feature as a risk factor and elevating a prior diagnosis over the true causal exposure. A cold start provides stronger initialization (improved terminology and patterns), but without CRPO’s constraints it does not consistently prevent these errors. With CRPO, reasoning remains faithful to the case, systematic in rejecting distractors, and anchored to explicit, verifiable evidence.

\subsection{Medical Faithfulness}
\label{sec:faithfulness}
We evaluate the factual and clinical quality of generated reasoning using four dimensions: faithfulness, case-grounded evidence citation density (CECD), distractor rejection coverage (DRC), and hallucination. Together these metrics reflect whether a model reasons accurately, cites case-relevant evidence, rejects incorrect hypotheses, and minimizes unsupported statements.

As shown in Table~\ref{tab:result}, CRPO consistently yields the strongest medical faithfulness across datasets. The improvements are most pronounced in CECD and DRC, where the CRPO-trained system produces explanations that repeatedly point back to concrete tokens and provide concise case-specific refutations. Clinical-R1-3B (cold start + CRPO) attains the best overall profile, indicating that verifiable, multi-objective rewards are essential for aligning reasoning with patient-specific evidence and medical standards.

The qualitative contrast is clear. Under GRPO, answers often rely on broad associations without citing the quoted findings, overlooking anchors such as “every 45 days” and “160 mg/dL,” or drifting into taxonomic imprecision like calling ovarian hyperthecosis a subtype to be used interchangeably. In another example, GRPO focuses on a background label (“previous breast cancer”) instead of the causal exposure explicitly present in the history (“previous radiation therapy”), despite pathologic cues like “centroblastic and immunoblastic cell presence.” In contrast, CRPO formats \texttt{<dx>} as a ledger of case facts and requires \texttt{<conclusion>} to cite them. Distractors are then rejected with brief, evidence-bound rationales.

Hallucination behavior follows the same pattern. Cold start, by itself, encourages longer but less organized answer and can overgeneralize, occasionally introducing unsupported claims or loosened taxonomy. CRPO suppresses this tendency by making unsupported additions costly: conclusions must point to sentences already established in \texttt{<dx>}, which reduces language drift and discourages invented labs, staging, or etiologies. GRPO alone reduces some errors tied to final-answer pressure but still permits generic dismissals and ungrounded phrasing when explicit evidence links are not rewarded.

In summary, cold start strengthens the initializer, but CRPO regulates medical faithfulness. Verifiable, multi-objective rewards convert raw capacity into an auditable workflow that consistently ties claims to quoted case facts, broadens case-grounded citations, expands targeted distractor rejection, and reduces unsupported content. Their combination in Clinical-R1-3B produces reasoning that is faithful to medical knowledge and reliably anchored to the patient at hand.

\subsection{Response Length Analysis}
\label{sec:length}
\begin{figure}
\centering
\includegraphics[width=1\linewidth]{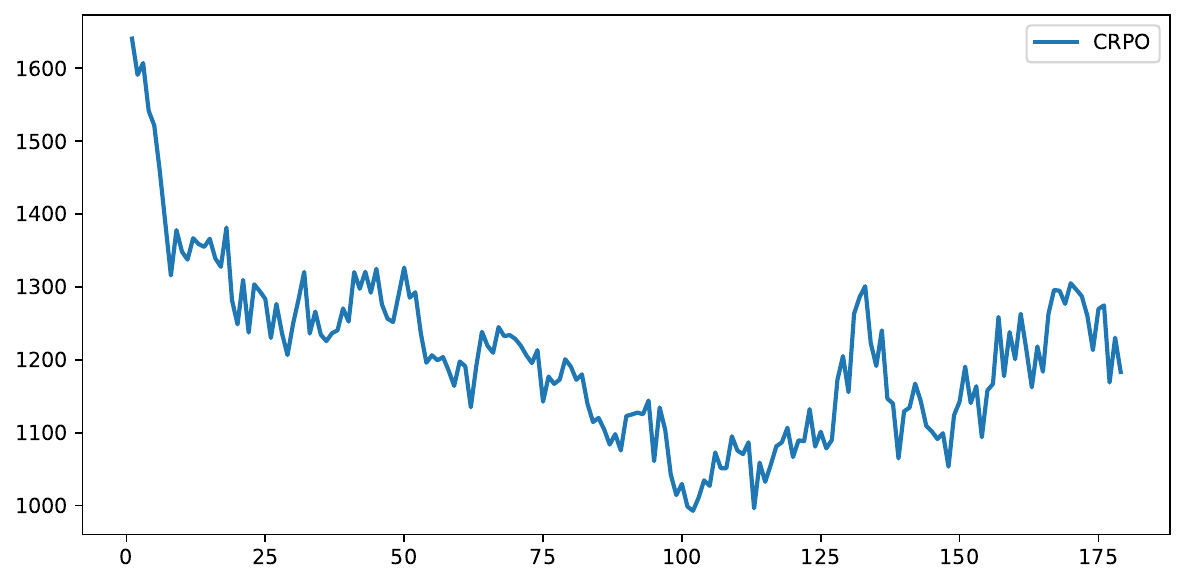}
\captionof{figure}{Response length vs training step (MedQA).}
\label{fig:response_length}
\end{figure}
We observe a characteristic length pattern across training stages. Cold start produces long, fully enumerated chains. GRPO then aggressively compresses responses. CRPO settles in between, yielding concise yet sufficiently structured explanations. This reflects CRPO’s ability to prune redundancy while preserving clinically necessary steps.

The trend is visible both in the stabilization of token counts during CRPO training (Figure~\ref{fig:response_length}). Superficial segments are progressively removed while essential reasoning is retained, indicating a shift from verbose exploration to compact reasoning.

\begin{figure*}[t]
\centering
\adjustbox{max width=\textwidth, max height=\textheight}{%
\begin{minipage}{\textwidth}

\begin{tcolorbox}[
  colframe=black,colback=white,boxrule=0.6pt,arc=6pt,
  width=1\linewidth,left=2mm,right=2mm,top=1mm,bottom=1mm,
]
\textbf{Question:}

{\footnotesize
A 77-year-old woman presents with fever, malaise, and night sweats. She recently noted enlargement of axillary lymph nodes that she examines weekly… She has a remote history of breast cancer in her 60s treated with radiation and chemotherapy; extensive travel to Africa; and a 30 pack-year smoking history. On exam, several axillary nodes are palpable with a large, non-tender right axillary mass (10$\times$8 cm). FNAB shows ``centroblastic and immunoblastic cell presence, suspicious for NHL--DLBCL variant.''\\
\textbf{Which risk factor is responsible?}\\
A) Travel to Africa \quad
B) Axillary lymph node involvement \quad
C) Previous radiation therapy \quad
D) Female gender \quad
E) Previous breast cancer \\[4pt]
\textbf{\textcolor{olive}{Groundtruth: C}}
}
\end{tcolorbox}

\begin{tcolorbox}[
  colframe=black,colback=white,boxrule=0.6pt,arc=6pt,
  width=1\linewidth,left=2mm,right=2mm,top=1mm,bottom=1mm,
]
\textbf{Cold Start + GRPO Answer}

{\footnotesize
\textless think\textgreater
``…77-year-old woman with fever, malaise, night sweats… enlarged axillary lymph nodes… FNA suggests DLBCL.'' \\
``Option A: Travel to Africa… but radiation therapy is a known risk factor…'' \\
``Option B: part of the presentation, not a risk.'' \\
``Option C: Previous radiation therapy… a big risk factor.'' \\
``Option E: Previous breast cancer… known risk factor for lymphoma…'' \\
\emph{(weighs options, then conflates prior disease with causation)} \textit{(…)}
\textless/think\textgreater

\textless answer\textgreater
Discusses prior breast cancer treated with radiation, but concludes \textbf{E (previous breast cancer)} as the ``risk factor'' rather than the treatment exposure. \\
\boxed{E}
\textless/answer\textgreater
}
\end{tcolorbox}

\begin{tcolorbox}[
  colframe=black,colback=white,boxrule=0.6pt,arc=6pt,
  width=1\linewidth,left=2mm,right=2mm,top=1mm,bottom=1mm,
]
\textbf{Cold Start + CRPO Answer}

{\footnotesize
\textless dx\textgreater
\textbf{(System 1: Intuitive)} ``…Age 77 with B-symptoms (fever, night sweats, malaise) and a bulky 10$\times$8\,cm axillary mass; FNAB ``centroblastic/immunoblastic'' $\rightarrow$ \textit{DLBCL plausible}.'' Notes history of breast cancer \emph{treated with radiation/chemotherapy}, Africa travel, and smoking… Key risk candidates arise: \textit{prior radiation}, immunosuppression/HIV/EBV, certain infections; gender is unlikely… \\[3pt]

\textbf{(System 2: Analytical)} \textit{Risk vs.\ finding clarification}: (B) \textit{axillary involvement} = presentation, not a risk. \\
(A) \textit{Travel to Africa}: can relate to endemic EBV/Burkitt… but \emph{not the pertinent causal risk} for this cytology. \\
(D) \textit{Female gender}: not established as a DLBCL risk. \\
(E) \textit{Previous breast cancer}: the disease history itself is \emph{not} causal; the \textbf{treatment exposure} is. \\
(C) \textbf{Previous radiation therapy}: consistent with radiation-induced lymphomagenesis (DNA damage, secondary NHL), matches this patient’s prior breast-cancer \emph{radiation}… \\[2pt]
Synthesizes: distinguish presentation vs.\ etiology; map each option to risk type; elevate \textbf{C} as the causal exposure while demoting non-causal context (A/E) and mere findings (B)… \textit{(…)}
\textless/dx\textgreater

\textless conclusion\textgreater
Most relevant risk factor for this DLBCL presentation is the patient’s \textbf{previous radiation therapy} (treatment exposure), not prior malignancy itself. \\
\boxed{C}
\textless/conclusion\textgreater
}
\end{tcolorbox}
\end{minipage}
}
\caption{Case study on a DLBCL risk-factor question. Ellipses (\dots) indicate trimmed content.}
\label{fig:case_study}
\end{figure*}

\subsection{Case Study}
\label{sec:case}
Figure~\ref{fig:case_study} shows a case example for cold-start + GRPO and cold-start + CRPO. For this particular case, in GRPO, the model frequently conflates disease labels with causal exposures, jumping from ``prior breast cancer'' to secondary malignancy risk without separating the disease label from the treatment exposure. Distractors such as sex, anatomic site, or travel are rarely preemptively negated, and mis-selection of ``previous breast cancer'' is common.

In Cold-start + GRPO, the model more often outlines alternatives and attempts eliminations, reflecting inherited structure. However, the final mapping remains vulnerable to high-salience yet non-causal labels; radiation exposure is not consistently isolated from the disease label. Errors decrease relative to pure GRPO but persist under strong distractors.

In CRPO, a three-way partition is established before scoring options: causal exposure (radiation) versus disease label/comorbidity (breast cancer) versus presentation/epidemiologic context (axillary involvement, sex, travel). The exposure timeline and directionality are aligned to the outcome, and non-causal features are proactively rejected. The model selects previous radiation therapy and provides explicit negative evidence for alternatives.

In Cold-start + CRPO, subgoals are articulated (categorize options, align case facts, eliminate distractors), and brief verification confirms that exposure precedes outcome. The resulting chain is compact and auditable: radiation is the causal exposure; breast cancer is a label; sex/site/travel constitute presentation or weak-context features. This case illustrates the general conclusion that cold start improves structural coverage, while CRPO, with or without cold start, provides stronger cognitive comprehensiveness and medical faithfulness.

\section{Conclusion and Limitation}
In this work, we presented Clinical-objective Relative Policy Optimization (CRPO), a scalable, multi-objective, and verifiable RL framework that aligns post-training with clinical reasoning principles. By combining rule-based rewards for accuracy, faithfulness, and comprehensiveness to enforce a lightweight reasoning answer (\texttt{<dx>}, \texttt{<conclusion>}). CRPO improves the quality of the reasoning process without relying on human annotation. Trained with CRPO, our 3B-parameter Clinical-R1-3B achieves stronger medical faithfulness and reasoning comprehensiveness than GRPO while maintaining higher answer accuracy across three benchmarks MedQA, MedMCQA, and MedXpertQA (Table~\ref{tab:result}). These gains are accompanied by desirable cognitive behaviors (backtracking, backward-chaining, subgoal setting, and verification), suggesting that verifiable, multi-objective optimization is an effective pathway for safer, more collaborative clinical LLMs.

Despite these promising results, several limitations remain. First, CRPO optimization can be unstable, particularly without a cold-start initialization, due to rapid KL-divergence growth between the policy and reference models. This instability may lead to inefficient training or partial convergence. Second, cognitive comprehensive and medical faithfulness evaluation rely on automatic annotation by another LLM, which may not fully align with human judgment. As a result, human evaluation is required for stronger validation. Third, our experiments are conducted on base models not pretrained on medical corpora, limiting the representational depth of domain knowledge.

Future work should explore more stable variants of CRPO (e.g., adaptive or off-policy updates), introduce human-in-the-loop evaluations, and extend the method to stronger medical backbones. We hope this line of research contributes toward building smaller yet more interpretable and trustworthy medical reasoning systems.

\section*{Acknowledgements}
DAC was funded by an NIHR Research Professorship; a Royal Academy of Engineering Research Chair; and the InnoHK Hong Kong Centre for Cerebro-cardiovascular Engineering (COCHE); and was supported by the National Institute for Health Research (NIHR) Oxford Biomedical Research Centre (BRC) and the Pandemic Sciences Institute at the University of Oxford. The Applied Digital Health (ADH) group at the Nuffield Department of Primary Care Health Sciences is supported by the National Institute for Health and Care Research (NIHR) Applied Research Collaboration Oxford and Thames Valley at Oxford Health NHS Foundation Trust. The views expressed are those of the author(s) and not necessarily those of the NHS, the NIHR or the Department of Health and Social Care. FL was funded by the Clarendon Fund and the Magdalen Graduate Scholarship. HZ was funded by the Clarendon Fund, the Department of Engineering Science Studentship, and the Frederick Brodckhues Scholarship. BMS is funded by the Rhodes Trust under the Rhodes scholarship. We sincerely thank all the editors and reviewers for their constructive comments and suggestions that substantially improved this paper.

\bibliography{refs,refs1_ext,refs2_ext,refs3_ext}

\end{document}